\documentclass[aip, 
jcp, longbibliography, 
twocolumn, floatfix, superscriptaddress, reprint, 10pt]{revtex4-1}
\usepackage[english]{babel}
\usepackage[utf8]{inputenc}
\usepackage[paperwidth=210mm,paperheight=297mm,centering,hmargin=2cm,vmargin=2.5cm]{geometry}
\usepackage[utf8]{inputenc}
\usepackage{verbatim}
\usepackage{comment}
\usepackage{bm}
\usepackage{amsmath}
\usepackage{amsfonts}
\usepackage{natbib}
\usepackage{pdfpages}
\usepackage{graphics}
\usepackage{float}
\usepackage{hyperref}
\usepackage[capitalize]{cleveref}
\usepackage{multirow}
\usepackage{booktabs}
\usepackage{makecell}
\usepackage{outlines}
\usepackage{lipsum}
\usepackage{siunitx}
\usepackage{mhchem}
\usepackage{caption}
\usepackage{subcaption}
\usepackage{xcolor}
\usepackage{ulem}
\usepackage{physics}
\usepackage{xparse}

\newcommand{\mbf}[1]{\ensuremath{\mathbf{#1}}}

\NewDocumentCommand{\rep}{s d<| d|>}{%
\IfBooleanTF{#1}{
   \IfValueTF{#2}{
       \IfValueTF{#3}{\braket{#2}{#3}}{\bra{#2}}
       }{
       \IfValueTF{#3}{\ket{#3}}{}
       }
   }{
   \IfValueTF{#2}{
       \IfValueTF{#3}{\braket*{#2}{#3}}{\bra*{#2}}
       }{
       \IfValueTF{#3}{\ket*{#3}}{}
       }
   }
}

\NewDocumentCommand{\rbra}{sm}{\IfBooleanTF{#1}{\rep*<#2|}{\rep<#2|}}
\NewDocumentCommand{\rket}{sm}{\IfBooleanTF{#1}{\rep*|#2>}{\rep|#2>}}
\NewDocumentCommand{\rbraket}{smom}{
    \IfBooleanTF{#1}{
        \IfNoValueTF{#3}{\rep*<#2||#4>}{\rep*<#2|#3\rep*|#4>}
    }{
        \IfNoValueTF{#3}{\rep<#2||#4>}{\rep<#2|#3\rep|#4>}
    }
}

\NewDocumentCommand{\cg}{m m m}{\rep<#1; #2||#3>}

\NewDocumentCommand{\field}{o m e{_} e{^} o e{_} e{^}}{
\IfValueTF{#5}{\overline{
  #2\IfValueT{#3}{_#3}\IfValueT{#4}{^{\otimes #4}} %
  \otimes
  #5\IfValueT{#6}{_#6}\IfValueT{#7}{^{\otimes #7}} %
  \IfValueT{#1}{;#1}
}}{
  \IfValueTF{#4}{\overline{
     #2\IfValueT{#3}{_#3}\IfValueT{#4}{^{\otimes #4}}
     \IfValueT{#1}{;#1}
  }}
  {#2\IfValueT{#3}{_#3}}
}
}

\NewDocumentCommand{\frho}{o e{_} e{^}}{
\field[#1]{\rho}_{#2}^{#3}
}

\newcommand{\e}{a}  %

\newcommand{\br}{\mbf{r}}
\newcommand{\bx}{\mbf{x}}
\newcommand{\bxhat}{\hat{\mbf{x}}}
\newcommand{\brhat}{\hat{\mbf{r}}}

\NewDocumentCommand{\ex}{e_}{
\IfValueTF{#1}{\e_{#1}\bx_{#1}}{\e\bx}
}  %

\NewDocumentCommand{\lm}{e_}{
\IfValueTF{#1}{l_{#1}m_{#1}}{lm}
}
\NewDocumentCommand{\nlm}{e_}{
\IfValueTF{#1}{n_{#1}\lm_{#1}}{n\lm}
}
\NewDocumentCommand{\enlm}{e_}{
\IfValueTF{#1}{\e_{#1}\nlm_{#1}}{\e\nlm}
}
\NewDocumentCommand{\en}{e_}{
\IfValueTF{#1}{\e_{#1}n_{#1}}{\e n}
}
\NewDocumentCommand{\nlk}{e_}{
\IfValueTF{#1}{n_{#1}l_{#1}k_{#1}}{nlk}
}
\NewDocumentCommand{\enlk}{e_}{
\IfValueTF{#1}{\e_{#1}\nlk_{#1}}{\e\nlk}
}
\NewDocumentCommand{\enl}{e_}{
\IfValueTF{#1}{\en_{#1}l_#1}{\en l}
}

\NewDocumentCommand{\nnl}{s}{
\IfBooleanTF{#1}{n_1 n_2 l}{n_1; n_2; l}
}
\NewDocumentCommand{\ennl}{s}{
\IfBooleanTF{#1}{\en_1 \en_2 l}{\en_1; \en_2; l}
}

\NewDocumentCommand{\gslm}{s}{
\IfBooleanTF{#1}{\sigma\lambda\mu}{\sigma;\lambda\mu}
}
\NewDocumentCommand{\glm}{}{\lambda\mu}

\newcommand{\bC}{\mbf{C}}
\newcommand{\bU}{\mbf{U}}
\newcommand{\bLam}{\boldsymbol{\Lambda}}

\newcommand{\mc}[1]{{\color{blue}{ #1}}}

\newcommand{\fm}[1]{{\color{purple}{ #1}}}

\newcommand{\lmax}[0]{l_\mathrm{max} }
\newcommand{\nmax}[0]{n_\mathrm{max} }
\newcommand{\qmax}[0]{q_\mathrm{max} }

\newcommand{\nsoap}[0]{{n_\mathrm{SOAP}} }

\newcommand{\fcut}[0]{{f_\mathrm{cut}} }
\newcommand{\rcut}[0]{{r_\mathrm{cut}} }

\newcommand{\D}[2][]{\ensuremath{\mathop{}\!\mathrm{d}^{#1}{#2}\,}}

\newcommand{\opt}{\text{opt}}

\newcommand{\alexnote}[1]{{\color{teal} #1}}
\definecolor{royalazure}{rgb}{0.0, 0.22, 0.66}
\newcommand{\revtwo}[1]{#1}
\newcommand{\rev}[1]{#1}
\newcommand{\rascal}[0]{{librascal}}
\newcommand{\todorev}[1]{{}}
\makeatletter
\AtBeginDocument{\let\LS@rot\@undefined}
\makeatother
\begin{document}

\setcitestyle{super}

\title{Optimal radial basis for density-based atomic representations}

\author{Alexander Goscinski}
\author{F\'elix Musil}
\author{Sergey Pozdnyakov}
\author{Jigyasa Nigam}
\author{Michele Ceriotti}
\email{michele.ceriotti@epfl.ch}
\affiliation{Laboratory of Computational Science and Modeling, Institute of Materials, \'Ecole Polytechnique F\'ed\'erale de Lausanne, 1015 Lausanne, Switzerland}

\onecolumngrid
\begin{abstract}

The input of almost every machine learning algorithm targeting the properties of matter at the atomic scale involves a transformation of the list of Cartesian atomic coordinates into a more symmetric representation. 
Many of the most popular representations can be seen as an expansion of the symmetrized correlations of the atom density, and differ mainly by the choice of basis. 
\rev{Considerable effort has been dedicated to the optimization of the basis set, typically driven by heuristic considerations on the behavior of the regression target. 
Here we take a different, unsupervised viewpoint, aiming to determine the basis that encodes in the most compact way possible the structural information that is relevant for the dataset at hand.
For each training dataset and number of basis functions, one can determine a unique basis that is optimal in this sense,} and can be computed at no additional cost with respect to the primitive basis by approximating it with splines. 
We demonstrate that this construction yields representations that are accurate and computationally efficient, \rev{particularly when constructing representations that correspond to high-body order correlations.} We present examples that involve both molecular and condensed-phase machine-learning models.

\end{abstract}
\twocolumngrid

\maketitle

\todorev{AUTHOR COMMENT KEY

\maxnote{Max Veit}

\fm{Félix Musil}

\alexnote{Alexander Goscinski}

\sp{Sergey Pozdnyakov}

\mc{Michele Ceriotti}
}

\section{Introduction}
Machine-learning algorithms for atomistic simulations rely heavily on the transformation of structural information and chemical composition into descriptors, or features.\cite{behl11pccp,rupp+12prl,bart+13prb} 
An effective molecular representation should be invariant (or more generally, equivariant) with respect to symmetry operations,\cite{behl-parr07prl,braa-bowm09irpc,bart+13prb,shap16mms,glie+17prb,gris+18prl,ande+19nips} be capable of differentiating between inequivalent configurations,\cite{pozd+20prl} and sensitive to atomic deformations\cite{onat+20jcp,pars+20mlst}. 
In broad terms it should encode in the most efficient way the relationships between a structure and the properties one is interested in predicting.\cite{huan-vonl16jcp} 
Even though many alternative approaches have been proposed to construct a representation that fulfils (at least partly) these requirements,\cite{musi+21cr} it has become clear that most of the existing schemes are strongly connected to each other, and can be seen as projections on different choices of basis functions of the correlations of the atom density~\cite{will+18pccp,will+19jcp}, or equivalently of a cluster expansion of interactions\cite{sanc+84pa,drau19prb}.
\rev{Besides the importance of these considerations to determine the formal relation between different frameworks, the choice of basis function affect the prediction quality\cite{zuo+20jpcl} and the efficiency of a basis in terms of linearly decodable mutual information\cite{gosc+21mlst}.
Consequently, several algorithmic recipes for the construction of basis  have been proposed\cite{shap16mms, drau19prb, bachmayr2019atomic, musi+21jcp} that aim at achieving computational efficiency, and/or at being best adapted to the specific requirement of a given fitting problem, typically the construction of a machine-learning model of the potential energy.
}
We bring these considerations to their logical conclusion, by showing that a data-driven basis to expand the atom density, \rev{that is optimal in terms of the information content for a given number of functions,} can be built as a contraction of a larger primitive basis set, similarly to what is routinely done in quantum chemistry for Gaussian type orbitals (GTOs)\cite{scha+92jcp}, and that it can be practically, and inexpensively, evaluated as a numerical basis with striking similarities to ideas in electronic-structure methods\cite{blum+13cpc}.
\rev{Using an effective basis reduces the number of features that are needed to encode the same information, and thereby reduces the training and prediction time of the resulting machine learning (ML) models.}
We demonstrate the accuracy, and the computational efficiency, of this approach for both the construction of machine-learning potentials for materials, and for the prediction of molecular properties. 

\section{Theory}

We use the bra-ket notation originally introduced in Refs.~\citenum{will+18pccp,will+19jcp}, and discussed in detail in Ref.~\citenum{musi+21cr}. 
An atomic structure $A$ is represented in terms of its atom density
\begin{equation}
\rep<\ex||A; \rho> = \sum_i \delta_{\e\e_i} \rep<\bx||\br_i; g>, 
\end{equation}
where $\rep<\bx||\br_i; g>\equiv g(\bx-\br_i)$, is a Gaussian of width $\sigma_a$ centered on the position $\br_i$ of the $i$-th atom, and $\e_i$ is an index that indicate the chemical species of that atom. 
Translational symmetrization breaks this global atom density into a sum of atom-centred neighbour densities $\rep|A; {\left<\rho^{\otimes 2}\right>_{\mathbb{R}^3}}>=\sum_i \rep|A; \rho_i>$, 
\begin{equation}
\rep<\ex||A; \rho_i> = \sum_{j\in A} \delta_{\e\e_j} \rep<\bx||\br_{ji}; g> \fcut(r_{ji}), 
\end{equation}
where $\br_{ji}=\br_j-\br_i$, and we  introduce a smooth cutoff function $\fcut$ to restrict the range of the environment.

It is convenient to express $\rep<\ex||A; \rho_i>$ on a basis of spherical harmonics $ Y^m_l(\hat{\bx})\equiv \rep<\bxhat||lm>$ and radial functions $R_{nl}(x) \equiv\rep<x||nl>$,
\begin{equation}
\rep<\enlm||A; \rho_i> =
\int \D{\bx} \rep<nl||x> \rep<lm||\hat{\bx}> \rep<\ex||A; \rho_i>.
\end{equation}
Regardless of the choice of $\rep<x||nl>$, one can evaluate the density coefficients as a sum over neighbors
\begin{multline}
\rep<\enlm||\rho_i> = 
\sum_j \delta_{\e\e_j} \rep<nlm||\br_{ji}; g> \\=
\sum_j \delta_{\e\e_j} 
\rep<nl||r_{ji}; g> \rep<lm||\brhat_{ji}>,
\label{eq:neighbor-sum}
\end{multline}
where $\rep<nl||r; g> $ is a radial integral
\begin{equation}
\!\!\!\rep<nl||r;g>= 4\pi e^{\!-\frac{r^2}{2\sigma_\e^2}}\!\! \int_0^\infty \!\!\!\D{x} x^2 \rep<nl||x> e^{\!-\frac{x^2}{2\sigma_\e^2}}\mathsf{i}_{l}\left(\frac{x r}{\sigma_\e^2}\right),
\label{eq:radial-integral}
\end{equation}
that can be computed analytically for some choices of basis, or approximated numerically and computed as a spline for each radial and angular channel pair\cite{musi+21jcp}.
The $\sigma_\e\rightarrow 0$ limit corresponds to a $\delta$-like density, which is used in alternative implementations of the density correlation features\cite{drau19prb,bachmayr2019arxiv,shap16mms}, and can be evaluated as easily on any discrete basis. 
We discuss in the SI some considerations on the practical evaluation of density coefficients. 

\subsection{Optimal density basis}

Principal component analysis has been proposed to compute the data-driven contractions of equivariant features that represent in the most informative way the variability of a dataset as part of the N-body iterative contraction of equivariant (NICE) frameworks.~\cite{niga+20jcp}
We propose to apply this procedure on the first-order equivariants -- that correspond to the density coefficients -- as a mean to determine a data-driven radial basis. 
Keeping different chemical species separate, this amounts to computing the rotationally invariant covariance matrix (see SI) 
\begin{equation}
C^{\e l}_{nn'}= \frac{1}{N} \sum_i \sum_m \rep<\e n l m||\rho_i> \rep<\rho_i||\e n' lm>,
\label{eq:cov}
\end{equation}
\rev{where the summation over $m$ ensures that the covariance is independent of the orientation of structures in the dataset.}
For each $(\e, l)$ channel, one diagonalizes $\bC^{al}=\bU^{al} \bLam^{al} (\bU^{al})^T$, and computes the optimal coefficients
\begin{equation}
\rep<\e q lm; \opt||\rho_i> = \sum_n U^{\e l}_{qn} \rep<\e n lm||\rho_i>. \label{eq:contract}
\end{equation}
Note that we compute $\bC^{al}$ without centering the density coefficients. For $l>0$, the mean ought to be zero by symmetry (although it might not be for a finite dataset), and even for the totally symmetric, $l=0$ terms, density correlation features are usually computed in a way that is more consistent with the use of non-centered features. 

The number of contracted numerical coefficients $\qmax$ can be chosen inspecting the eigenvalues $\Lambda^{al}_q$.
At first, it might appear that in order to evaluate the contracted basis one has to compute the full set of $\nmax$ coefficients, and this is how the idea was applied in Ref.~\citenum{niga+20jcp}. 
When combining Eq.~\eqref{eq:contract} with Eq.~\eqref{eq:neighbor-sum}, however, one sees that the contracted coefficients can be evaluated directly 
\begin{equation}
\!\!\rep<\e qlm; \opt||\rho_i> \!=\! \sum_j \delta_{\e\e_j} \rep<\e ql; \opt||r_{ji}; g>\! \rep<lm||\brhat_{ji}>,
\label{eq:neighbor-contract}
\end{equation}
using the contracted radial integrals
\begin{equation}
\rep<\e q l; \opt||r; g> = \sum_n U^{\e l}_{qn} \rep<nl||r; g>, \label{eq:contract-integral}
\end{equation}
\rev{that can be computed over $r$, approximated with cubic splines in the range $[0, r_c]$}, and then evaluated at exactly the same cost as for a spline approximation of the radial integrals of a primitive basis of size $\qmax$.
\rev{The exact mathematical form and implementation details of the splines can be found in Ref. \citenum{musi+21jcp}.}
Splining does not affect the equivariant behavior of the atom-density features, and introduces minute discrepancies relative to the analytical basis that do not affect the quality of the resulting models. 
\rev{
Thus, the procedure we propose entails the following steps:
\begin{enumerate}
\item Compute the density coefficients \eqref{eq:neighbor-sum} for a representative dataset, using \emph{any} primitive basis, and a large $\nmax$
\item Compute the covariance~\eqref{eq:cov} and diagonalize it, finding the contraction coefficients $U^{\e l}_{qn}$
\item Evaluate the contracted radial integrals using Eq.~\eqref{eq:contract-integral}, over a dense radial grid
\item Use a spline approximation to evaluate directly the radial integrals~\eqref{eq:neighbor-contract} for the first $\qmax$ optimal features, and use the coefficients in subsequent ML steps
\end{enumerate}
}

\rev{
Even though this framework only needs the contracted radial \emph{integrals}~\eqref{eq:radial-integral}, one can also compute and inspect the ``optimal radial basis'' that corresponds to the optimized coefficients  
\begin{equation}
\rep<x||\e ql; \opt> \equiv \sum_n U^{\e l}_{qn} \rep<x||nl>.\label{eq:contract-basis}
\end{equation}
For a given dataset, these functions are optimal in the sense that when truncated to $\qmax<\nmax$, they describe the greatest fraction of the variance for the local atom-density coefficients, and unique in the sense that they are independent on the choice of the primitive basis, in the limit in which the latter is complete, as demonstrated in Sec.~\ref{sec:practice}.
}

\rev{
\paragraph*{Mixed-species basis}

Even though Eq.~\eqref{eq:cov} is defined separately for different species $\e$, it is also possible to compute cross-correlations between different elemental channels, defining
\begin{equation}
 C^{l}_{\e n; \e'n'}= \frac{1}{N} \sum_i \sum_m \rep<\e n l m||\rho_i> \rep<\rho_i||\e' n' lm>,
\label{eq:cov-multispecies}
\end{equation}
as done in the NICE framework\cite{niga+20jcp} following ideas proposed in Ref.~\citenum{will+19jcp}, resulting in coefficients that combine information on multiple species
\begin{equation}
\rep<q lm; \opt||\rho_i> = \sum_n U^{l}_{q;\e n} \rep<\e n lm||\rho_i>, \label{eq:contract-multispecies}
\end{equation}
similar in spirit to the alchemical contraction discussed in Ref.~\citenum{will+18pccp}.
It is worth noting that although the NICE code\cite{NICE-REPO} contains the infrastructure to compute these contractions \emph{as a post-processing of the primitive basis}, the implementation we propose in  \rascal{}\cite{LIBRASCAL} computes the contracted coefficients directly. However, it only implements the less information-efficient separate $(\e, n)$-PCA strategy.
An implementation that evaluates directly the combined contraction would incur an overhead because every neighbor would contribute to every $q$ channel irrespective of their species:
\begin{multline}
\rep<qlm; \opt||\rho_i>  = \sum_j \sum_{\e n} U^l_{q; \e n} \delta_{\e \e_j}
\rep<nlm||\br_{ji}; g>  \\
= \sum_j \sum_{n} U^l_{q; \e_j n} \rep<nl||r_{ji}; g>\rep<lm||\brhat_{ji}> \\=
\sum_j \rep<\e_j q l; \opt||r_{ji}; g>\rep<lm||\brhat_{ji}>.
\end{multline}
Given however that the cost of evaluating the density coefficients is usually a small part of the calculation of density-correlation features\cite{caro19prb,musi+21jcp}, we expect that this approach should be in general preferable compared to the calculation of a large primitive basis, and to a two-step procedure in which element-wise optimal functions are further contracted into mixed-element coefficients.

\paragraph*{Supervised basis set optimization}
For a given number of radial functions, and a target data set, the data-driven contracted basis~\eqref{eq:contract} provides the most efficient description of the atom-centred density in terms of the fraction of the retained variance.
The most effective variance-preserving compression however does not guarantee that the features are the most effective to predict a given target property.
In fact, it has already been shown that SOAP features tend to emphasize correlations between atoms that are far from the atomic center, which can lead to a counter-intuitive degradation of the model accuracy with increasing cutoff radius\cite{bart+17sa,will+18pccp}. 
This effect can be contrasted by introducing a radial scaling\cite{huan-vonl16jcp,will+18pccp} that de-emphasizes the magnitude of the atom density in the region far from the central atom. 
By applying this scaling -- or other analogous tweaks\cite{caro19prb} -- to the atom density before it is expanded in the primitive basis, one ensures that the optimal basis is also built with a similar focus on the structural features that contribute more strongly to the target property.
In other terms, the information-optimal basis set we introduce here can be combined with a heuristic or data-driven optimization of the underlying density representation, to reflect the scale and resolution of the target property.
}

\rev{Another possibility is to extend the scheme to incorporate a supervised target $y_i$ in the selection of the optimal basis using principal covariates regression (PCovR) \cite{dejo-kier92cils,helf+20mlst}.
PCovR is a simple linear scheme that can be tuned to provide a projection of features to a low-dimensional latent space that combines an optimal variance compression target with that of providing an accurate linear approximation of the desired target property. 
Since $l>0$ contributions of the features have zero mean, the optimization problem can be combined with a supervised component only for $l=0$, and yields an optimal basis
\begin{equation}
\rep<r||  \e q 0; \opt; \gamma> = \sum_n U^{\e0;\gamma}_{qn} \rep<r||n0>,
\end{equation}
which is a special case of Eq.~\eqref{eq:contract-basis} for $l=0$, 
where $U^{\e0;\gamma}_{qn}$ is obtained as the orthogonalized PCovR projector, as discussed in Refs.~\citenum{dejo-kier92cils,helf+20mlst}, using a mixing parameter $\gamma$, that determines how strong the emphasis of the optimization should be on minimizing the residual variance or the error in regressing the target.
}

\subsection{Density correlation features}
\label{sub:multispectrum}

In the vast majority of applications the density coefficients are not used directly in applications, but are combined to build higher order invariant or equivariant features\cite{bart+13prb,gris+18prl,will+19jcp,drau19prb}.
For example, the power spectrum (i.e. SOAP invariant features\cite{bart+13prb}) can be computed as
\begin{multline}
\rep<\ennl||\frho_i^2> \propto\\
\frac{1}{\sqrt{2l+1}} \sum_m\rep<\en_1 lm||\frho_i>\rep<\en_2 l m||\frho_i>^\star,
\end{multline}
where the density coefficients can be either those obtained from \rev{primitive basis} functions truncated at increasing $\nmax$, or those from an optimal basis containing $\qmax$ terms.
\rev{
For this work we use primarily the orthogonalized GTO basis introduced in Ref.~\citenum{musi+21jcp}, which compares favorably in terms of information content\cite{gosc+21mlst,musi+21cr} with a DVR basis (a family of orthogonal polynomials), as well as with the alternative GTO basis used in DScribe\cite{Himanen2020} and the shifted-Gaussian basis of QUIP\cite{bart+10prl}.
}

We discuss the general case of ``multispectra'' in the frame of the N-body iterative construction of equivariant (NICE) features\cite{niga+20jcp}, but analogous considerations apply to similar many-body descriptors such as the atomic cluster expansion (ACE)\cite{drau19prb,bachmayr2019arxiv} or the moment tensor potential (MTP)\cite{shap16mms}, and is likely to be relevant also for covariant neural networks\cite{ande+19nips,mill2020arxiv}. 
We consider the case of a single chemical species, to keep a notation that is by necessity quite cumbersome as simple as possible, but the generalization is trivial.
The NICE iteration increases the body order of features that describe correlations between $\nu$ neighbors $\rep<Q||\frho[\sigma; \lambda \mu]_i^{\nu}>$ ($Q$ is a generic index that labels the features, $\lambda$, $\mu$, $\sigma$ are indices that describe their behavior with respect to rotations and inversion) by combining lower order features
\begin{multline}
\rep<Q;\nlk||\frho[\gslm]_i^{(\nu+1)}>\propto 
\sum_{m}   \rep<n||\frho[\lm]_i^1>\times\\[-1mm]
\rep<Q||\frho[\sigma((-1)^{l+k+\lambda}); k (\mu-m)]_i^{\nu}>  \cg{\lm}{k (\mu-m)}{\glm},
\end{multline}
using Clebsch-Gordan coefficients $\cg{lm}{l'm'}{l''m''}$ in an expression analogous to the sum of angular momenta.
The $\nu=1$ equivariants are nothing but the density coefficients
\begin{equation}
\rep<n||\frho[\sigma;\lm]_i^1>  = \delta_{\sigma 1}\rep<\nlm||\rho_i>^\star,
\end{equation}
and one can compute invariant descriptors by retaining only the $\rep<Q||\frho[1; 00]_i^\nu>$ terms, using the other components only as computational intermediates.

\paragraph{Change of basis for the multi-spectrum}
First, we investigate the relation between the multispectrum computed in an arbitrary radial basis, and in the optimal basis obtained from the principal components of the density coefficients. 
\begin{multline}
\rep<Q;qlk; \text{opt}||\frho[\gslm]_i^{(\nu+1)}>
\propto \\
\sum_{m} 
\rep<qlm;\text{opt}||\frho_i>^\star 
\rep<Q||\frho[\sigma((-1)^{l+k+\lambda}); k (\mu-m)]_i^{\nu}> \\[-3mm]
\times \cg{\lm}{k (\mu-m)}{\glm}  \\[1mm]
=
\sum_{m} \cg{\lm}{k (\mu-m)}{\glm} \sum_n U^l_{qn} \rep<nlm||\frho_i>^\star  \\[-1mm] 
\times
\rep<Q||\frho[\sigma((-1)^{l+k+\lambda}); k (\mu-m)]_i^{\nu}>  \\[-1mm]
=\sum_n U^l_{qn}
\rep<Q;nlk||\frho[\gslm]_i^{(\nu+1)}>. \label{eq:multispectrum-transform}
\end{multline}
In other terms, the change of basis can be achieved by constructing the multispectrum using the density coefficients in the optimal radial basis, or by applying the transformation to each $(n_\nu,l_\nu)$ term in the multispectrum computed in the original basis. The transformation of the multi-spectrum is given by a block-diagonal matrix composed of the $\bU^{l}$. 

\paragraph{Truncation of the multispectrum}

Among the consequences of Eq.~\eqref{eq:multispectrum-transform} is the fact that -- if the optimal basis is not truncated, so that $\bU^{l}$ enacts an orthogonal transformation -- the change to the optimal basis preserves the magnitude of the multi-spectrum:
\begin{multline}
\sum_{q=1}^{\nmax}\left|\rep<Q;qlk; \text{opt}||\frho[\gslm]_i^{(\nu+1)}>\right|^2\\ =\sum_{n=1}^{\nmax} \left|\rep<Q;nlk||\frho[\gslm]_i^{(\nu+1)}>\right|^2.
\label{eq:multispectrum-norm}
\end{multline}

More generally, truncating the basis to include $\qmax$ optimized basis functions reduces the norm of the multispectrum by  the same multiplicative factor at each iteration
\begin{multline}
\sum_{q=1}^{\qmax}\sum_{lkQ}\sum_{\sigma\lambda\mu}\left|\rep<Q;qlk; \text{opt}||\frho[\gslm]_i^{(\nu+1)}>\right|^2
\\
\!\!\!=\sum_{q=1}^{\qmax }
\sum_{lm} \left|\rep<q||\frho[\lm]_i^1>\right|^2 \! \times \!\sum_{Q\sigma kp} \left|\rep<Q||\frho[\sigma; k p]_i^{\nu}>\right|^2 \label{eq:multi-truncation}
\end{multline}
which can be derived exploiting the orthogonality of CG coefficients (see SI).
One sees how (if the compound index $Q$ was expanded to indicate the $q_\nu l_\nu k_\nu$ terms at each  order $\nu$) the norm of the multispectrum can be expanded into a product of terms coming from each order, and the errors introduced by truncation accumulate as a product.
As a side-note, the combination of Eqs.~\eqref{eq:multispectrum-norm} and~\eqref{eq:multi-truncation} implies that, for each environment, the norm of the $\nu$-spectrum should equal the norm of the corresponding $1$-spectrum raised to the power $\nu$ \emph{when summing over all the equivariant components}. This provides a stringent test to estimate the amount of information that is lost when contracting, subselecting, or truncating the angular momentum of the equivariant components during the iterative construction of high body-order features.

\paragraph{Principal-component basis for multi-spectra}

The derivation of \eqref{eq:multi-truncation} applies to each environment $A_i$ separately, and does not translate exactly into an expression for the retained variance (that involves an average over the training set). 
A similar issue arises when addressing the question of what is the best radial basis (again, in terms of variance retained for a given level of truncation) that one can use to apply the NICE iteration for a specific feature $Q$ and intermediate angular momentum state $k$. 
In building the covariance, we sum over $(\sigma,\lambda,\mu)$ -- i.e. we look for a single transformation that applies to all terms that derive from combinations of $\rep<Q||\frho[s; k p]_i^{\nu}>$ with the density coefficients
\begin{multline}
N C^l_{nn'}(\nu;Q;k) = \\
\sum_{i\sigma\lambda\mu}
\rep<Q;\nlk||\frho[\gslm]_i^{(\nu+1)}>\rep<\frho[\gslm]_i^{(\nu+1)}||Q;n'lk>
\\=
\sum_i 
\sum_m \rep<n||\frho[\lm]_i^1> \rep<\frho[\lm]_i^1||n'> \\[-3mm] \times
\sum_{\sigma p} \left|\rep<Q||\frho[\sigma; k p]_i^{\nu}>\right|^2.
\end{multline}
This expression corresponds to a covariance matrix of the density coefficients which is built by weighting the contribution from each environment by the magnitude of $\rep<Q||\frho[\sigma; k p]_i^{\nu}>$. 
Thus, the optimal combinations that are determined for $\nu=1$ are not necessarily equal to those needed in further iterations. 
Computing a different radial basis for each NICE iteration would be extremely cumbersome; in what follows, we provide evidence that the basis optimized for the density coefficients provides an effective compression even for the higher-order terms in the multispectrum.

\begin{figure}[tbhp]
    \centering
    \includegraphics[width=\linewidth]{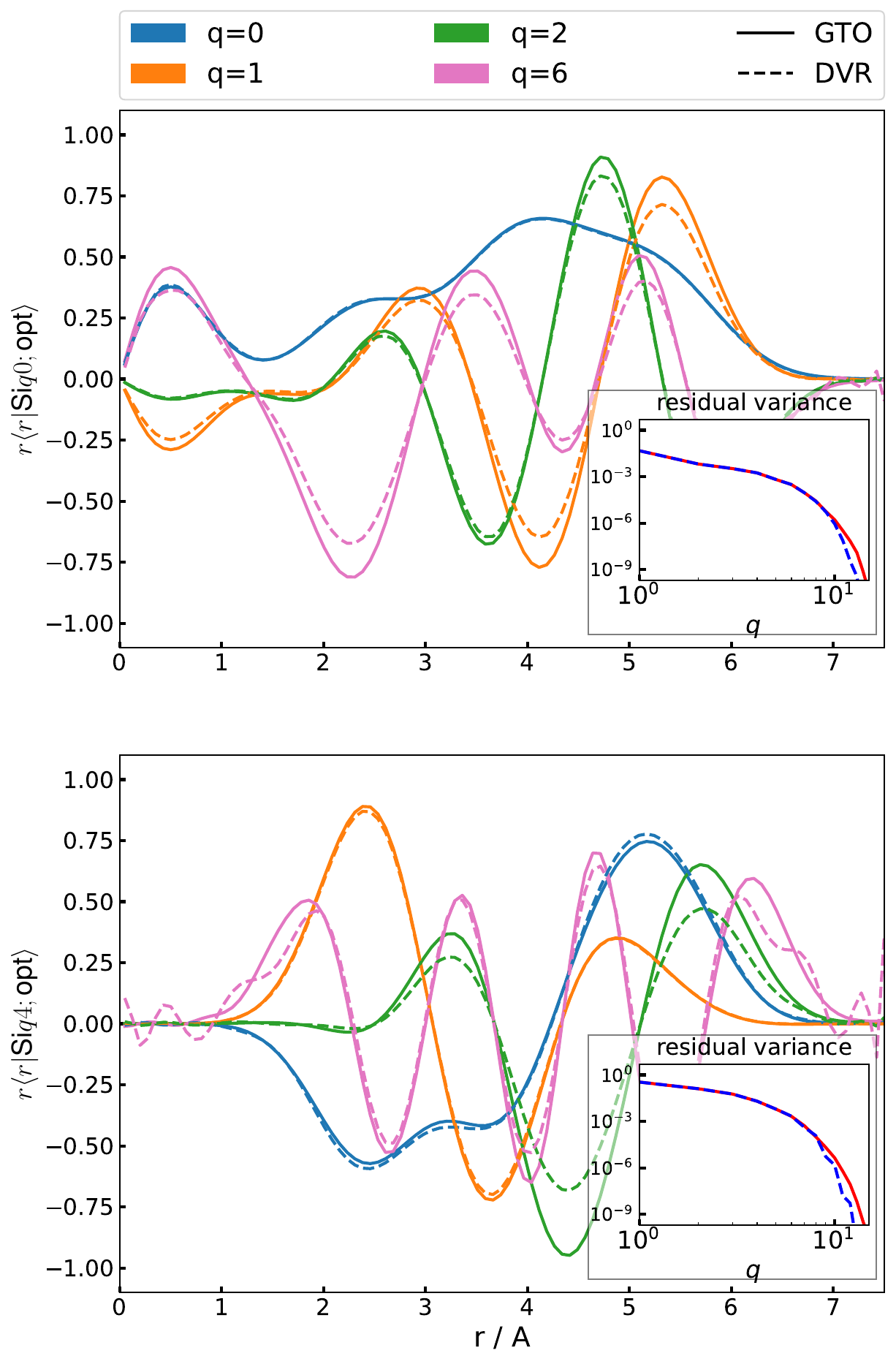}
    \caption{Several examples of the optimized radial basis functions on the silicon dataset for $l=0$ and $l=4$ using DVR and GTO as primitive basis contracted from $\nmax=20$, with $\rcut=6$. }%
    \label{fig:oel-radial-basis}
\end{figure}

\section{Results}
\label{sec:practice}
To illustrate the construction and use of an optimal radial basis we present examples for two very different problems: the construction of a general-purpose potential for silicon, based on the training dataset from Ref.~\citenum{Bartok2018}, and the prediction of atomization energies for the organic molecules from the QM9 dataset~\cite{rama+14sd}.
These two examples are complementary: the silicon potential involves a single chemical species, uses forces for training and aims to predict the properties of arbitrary distorted configurations.
The QM9 energy model involves multiple elements, but only minimum-energy structures, and, despite its limitations, has been widely used as a benchmark of new representations for molecular machine learning\cite{fabe+17jctc}.

\begin{figure}[tbhp]
    \centering
        \includegraphics[width=\linewidth]{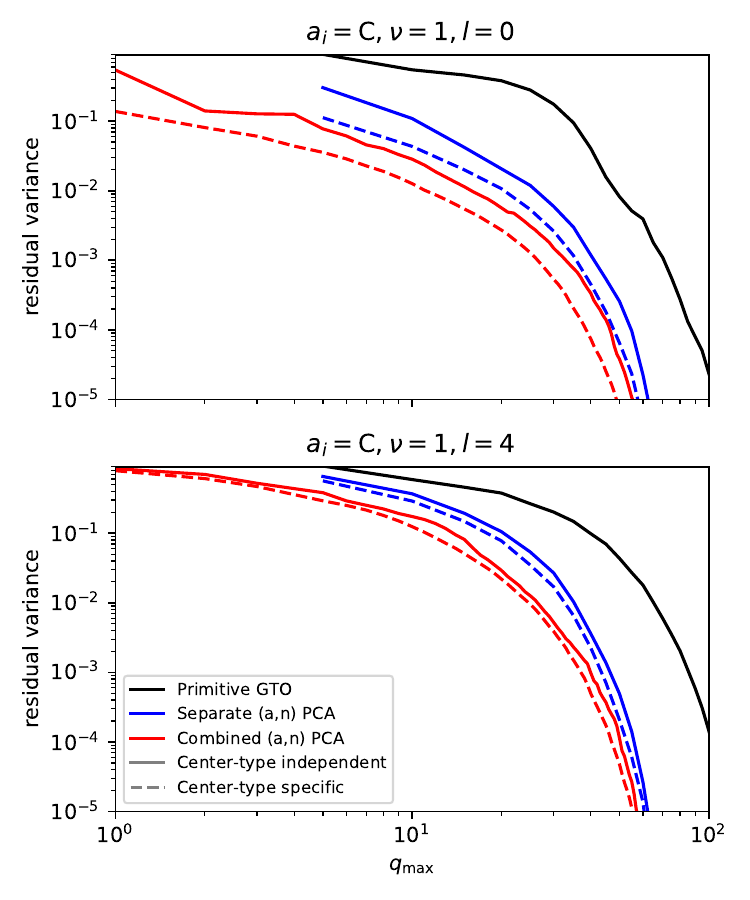}
    \caption{ Convergence of the residual variance for the expansion coefficients of the density as a function of the number radial basis functions $\qmax$, computed for the QM9 dataset and for environments centered on a C atom. The different series correspond to a GTO basis of increasing size (black), to an optimal basis computed for each neighbor density by separating (blue) or by mixing chemical and radial channels $(\e, n)$ (red). Full lines use the same basis irrespective of the species of the central atom, dashed lines correspond to a basis optimized specifically for C-centered environments.  }
    \label{fig:qm9-density-variance}
\end{figure}

\subsection{Convergence of the density expansion}

\rev{
We begin by considering the convergence of the density expansion, by considering a large primitive basis and then increasing $\qmax$ monitoring the residual variance
\begin{equation}
RV = 1-\frac{\sum_i\sum_{q=1}^{\qmax} \left|\rep<qlm; \opt||\rho_i> \right| ^2}{\sum_i\sum_{n=1}^{\nmax} \left|\rep<nlm||\rho_i> \right|^2},
\end{equation}
that measures the amount of information lost relative to that contained in the large-$\nmax$ primitive basis description. 
}
For the Si dataset, the residual variance decays rapidly with increasing number of optimal basis functions, as shown in Fig.~\ref{fig:oel-radial-basis}.
The figure also shows the shape of the optimal radial functions, and demonstrate that the same radial functions can be obtained starting from either of the DVR or GTO bases implemented in \rascal{}: the discrepancy increases for higher indices $q$, but can be reduced by increasing the size of the primitive basis, at no cost during the evaluation of the optimal splined basis. Furthermore, the optimal functions reflect some ``sensible'' expectations -- highly oscillating functions are associated with low covariance eigenvalues, the functions decay at the cutoff distance (even if the raw basis exhibits much larger spillover, see SI),
and higher angular momentum functions are peaked at larger distances, consistent with the greater variability in the angular distribution at large $r$.

In the multi-species case, exemplified by the QM9 dataset, there are several possible choices for the contraction strategy. First, one can compute a different contraction depending on the species of the central atom \rev{(center-type specfic)}, or use the same basis functions independent of $\e_i$ \rev{(center-type independent)}. Second, one can contract separately the density contribution from each neighbour type along the radial index, or compute a covariance matrix that combines the $(a,n)$ indices. 
Figure~\ref{fig:qm9-density-variance} shows the convergence of the explained variance for the four possible cases, compared to the baseline of a primitive GTO basis of increasing size - which shows by far the slowest convergence of the explained variance, requiring almost 100 radial channels ($\nmax=20$, for the 5 species present) to reduce the importance of features below $10^{-4}$.
The same level can be achieved with $\qmax\sim\; 50$ when performing separate PCAs for each neighbor species, and $\qmax\sim 30$ when computing jointly the correlations between radial and elemental channels. 
Performing a separate PCA depending on the species of the central atom accelerates slightly the convergence of the explained variance.

\subsection{Convergence of density correlations features}

We now turn to considering how the truncation of the density expansion basis affects the evaluation of higher-order features, focusing in particular on the invariant components.
We begin analyzing the convergence of the power spectrum computed for the Si dataset. 
We take the SOAP features computed with a large $\nmax=20$ as the ``full'' description of three-body correlations, and compute the global feature space reconstruction error\cite{gosc+21mlst} (GFRE) that measures how accurately the full feature space can be reconstructed using SOAP features that are built from a truncated density expansion. \rev{Given that SOAP features are usually subselected using a low-rank matrix approximation (CUR) approach\cite{imba+18jcp} and farthest point sampling (FPS)\cite{elda+97ieee,ceri+13jctc}, we also investigate the interplay between the density expansion optimization, and this further feature reduction step.}

\begin{figure}[bp]
    \centering
        \includegraphics[width=\linewidth]{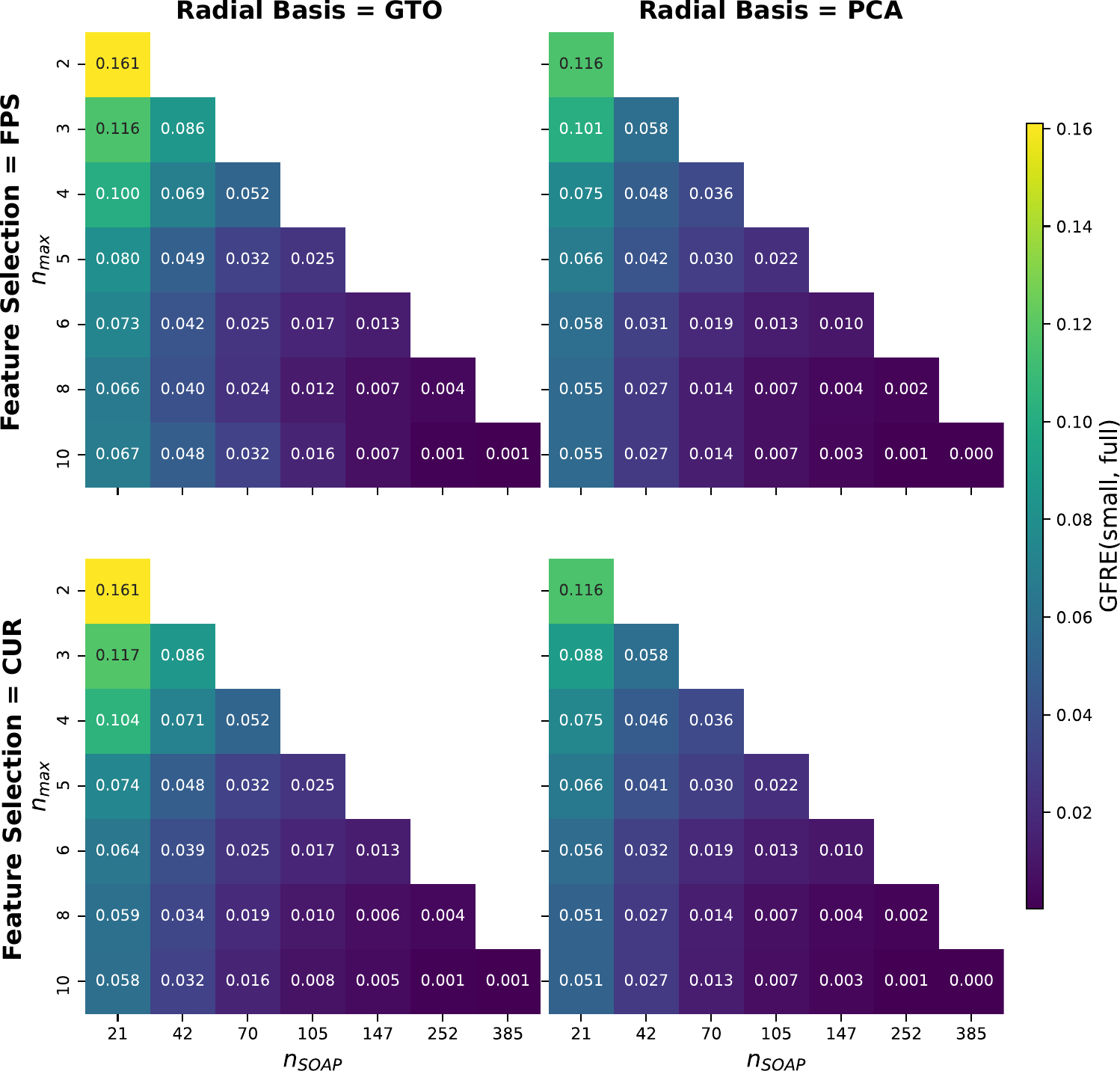}
    \caption{Feature space reconstruction errors for the power spectrum, resulting from the truncation of the radial basis and from the selection of a subset of the power spectrum entries using a deterministic CUR scheme and FPS. The ``full'' feature space is approximated with the power spectrum features, computed using a GTO basis with $(\nmax= 20, \lmax=6)$, and we compare the convergence obtained by using a smaller GTO basis against a truncated optimal basis of the same size.}
    \label{fig:silicon-soap-gfre}
\end{figure}

\begin{figure}[tbp]
    \centering
        \includegraphics[width=\linewidth]{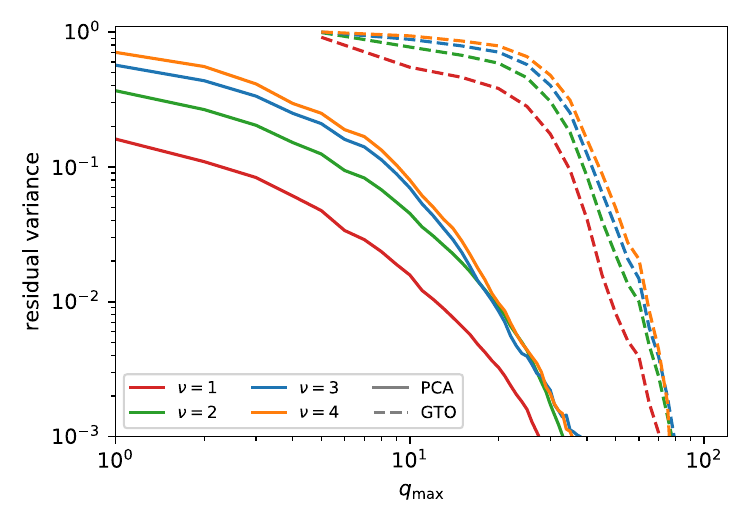}
        \includegraphics[width=\linewidth]{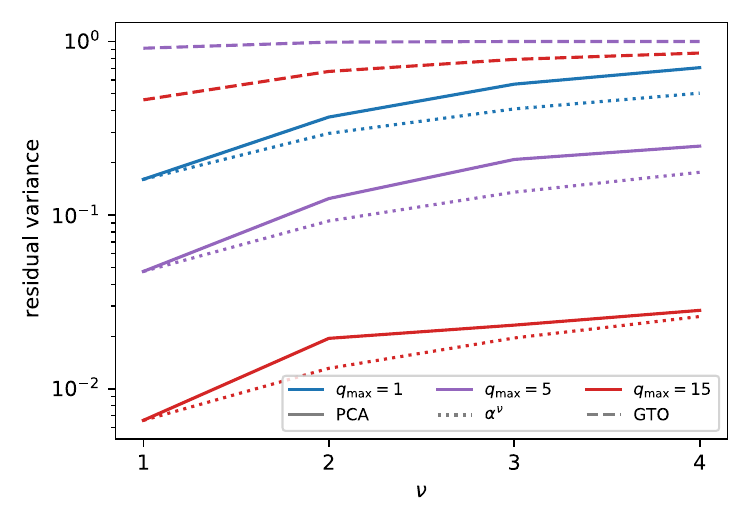}
    \caption{ Residual variance for the multispectra computed for the QM9 dataset. 
    For each body order, the baseline variance is taken to be that associated with the NICE features built starting from a ``full'' vector of density coefficients ($\nmax=20,  \lmax=5$) -- summing over the contributions from all atoms in a representative sample of the QM9 dataset. 
    We compare results for a small GTO basis (dashed lines) against those for an optimal basis (full lines) determined using a separate PCA procedure depending on the chemical nature of the central atom, and using a combined $(\e,n)$ covariance. 
(top) Different colors correspond to order-$\nu$ multispectra. $\nu=1$ and $\nu=2$ terms are computed in full; for the $\nu>2$ terms the NICE contraction has been converged so that the discarded variance at each iteration is smaller than that due to the truncation of the density coefficients. (bottom) Comparison of the residual variance for fixed radial/chemical basis size and different orders of multispectrum. Dotted lines indicate the behavior one would expect if the retained variance followed exactly the multiplicative behavior given in Eq.~\eqref{eq:multi-truncation}.}
    \label{fig:qm9-multispectrum-variance}
\end{figure}

Using an optimal density expansion basis systematically improves the GFRE compared to a GTO basis of the same size (Figure~\ref{fig:silicon-soap-gfre}).  
This is true both for the full-sized SOAP vector, and for a subselection of the invariant power spectrum entries based on a deterministic CUR algorithm, as well as on FPS.
This suggests that using an optimal radial basis as the building block of higher-order spectra yields feature vectors that can be easily compressed further, which is important to reduce the cost of evaluating SOAP based models.
\revtwo{ The cost of different parts of the feature evaluation (density expansion, invariant calculation, kernel evaluation, gradients ...)  depends subtly on the composition of the system and the various convergence parameters.\cite{musi+21jcp} When evaluating a Gaussian process regression model, the calculation of the invariant features and of the kernel values is often dominant, and so the possibility of aggressively subselecting SOAP features with little performance loss is as important as the reduction in the number of radial basis size. }

The same efficient compression is observed for the QM9 dataset, when extending the construction to higher-order features and to a multi-component system. Despite the fact that, as discussed in Section~\ref{sub:multispectrum}, there is no formal guarantee that the optimal density coefficients are also optimal to build high-$\nu$ equivariants,
we find in practice that the PCA basis leads to much faster convergence of the bispectrum and the trispectrum compared to the primitive basis (Fig.~\ref{fig:qm9-multispectrum-variance}, top panel). 
The truncation of the density coefficients affects the multispectra in a way that is qualitatively similar to what predicted by Eq.~\eqref{eq:multi-truncation}: the impact of an incomplete description of the density gets amplified by taking successive orders of correlation (Fig.~\ref{fig:qm9-multispectrum-variance}, bottom panel). 
\rev{Given that the raw number of multispectrum components grows exponentially as $\qmax^\nu$, the density basis truncation has a dramatic effect in reducing the size of the multispectrum vector.  This observation may be extremely important in the construction of systematic high-body order expansions such as NICE or ACE, and in particular in the extension of these approaches to multiple chemical species. }
The very efficient feature reduction that can be achieved by combining $(\e,n)$ channels at the density level shall make it much easier to avoid the exponential increase of complexity of high-body order models with growing chemical diversity. 

\subsection{Regression models}

\begin{figure}[b]
    \centering
        \includegraphics[width=\linewidth]{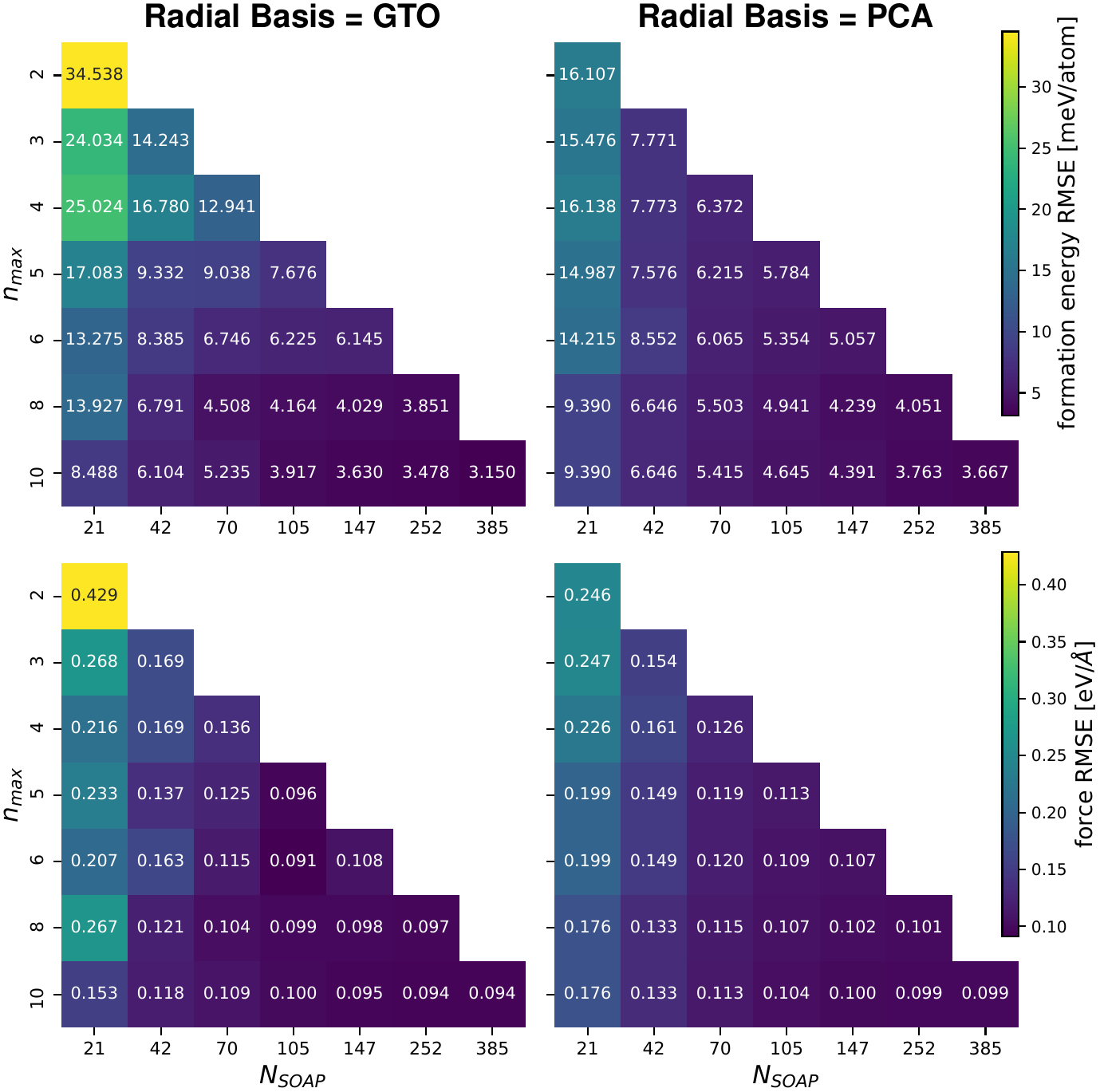}
    \caption{Energy and force RMSE for a Gaussian approximation potential based on the power spectrum, fitted to the Si dataset, plotted as a function of the number of radial functions $\nmax$($\qmax$) and sparsification of the SOAP features, $\nsoap$ (using CUR selection).}
    \label{fig:si-nmax_vs_feat-rs-rmse}
\end{figure}

The accuracy of a Gaussian approximation potential based on SOAP features, trained using both energy and forces (details in the SI), seen in Fig.~\ref{fig:si-nmax_vs_feat-rs-rmse} shows an improvement of the cross-validation error for the most aggressive truncation of the feature space (up to $\nmax\approx 6$ for forces, and $\nmax\approx 4$ for energy), but no improvements for large $\nmax$. 
For the largest feature set the primitive GTO basis can be up to 10\% more accurate than the corresponding optimal-basis model.
A comparison with Fig.~\ref{fig:silicon-soap-gfre}, that shows that the PCA basis is objectively more informative than the primitive basis, suggests that an effect similar to the degradation of performance with increasing environment cutoff radius might be at play here: for this dataset size, the GTO basis, which becomes smoother for large distances, is better suited to build a potential with limited amounts of training data.
The fact that the GTO basis may be fortuitously better adapted to this specific regression problem is also suggested by the non-monotonic convergence of the error. Depending on the value of $n_\text{max}$, the GTO functions are distributed so as to span the $[0, r_c]$ range (see SI).
Particularly for small $n_\text{max}$, the varying positions of maxima and nodes of the orthogonalized GTOs emphasize different portions of the atomic environment, and can produce such a non-monotonic trend, particularly in the limit of a relatively small train set size. 
The PCA basis, on the other hand, is constructed to provide a progressively more complete description of the atom density for the specific training set, resulting in a more regular, mostly monotonic convergence.

\begin{figure}[t]
    \centering
    \includegraphics[width=1.0\linewidth]{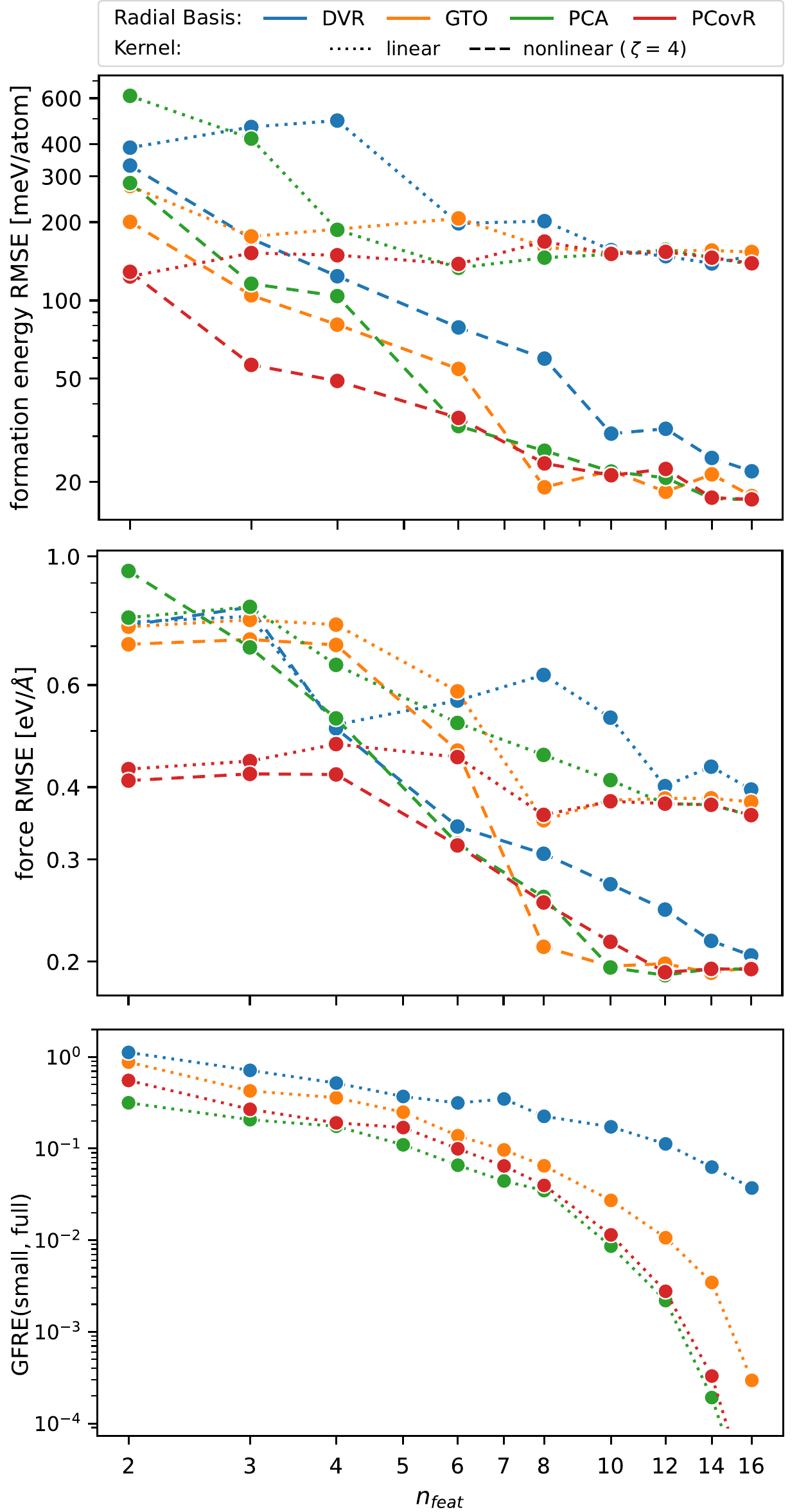}

    \caption{Energy (top) and force (center) 5-fold cross-validation RMSE and GFRE (bottom), computed on the silicon dataset for models based on the radial spectrum $\rep|\frho_i^1>$, as a function of the number of radial functions. Different curves correspond to a primitive DVR and GTO basis, and to the optimal (PCA and PCovR) contracted bases. The PCovR contraction is performed with $\gamma=0.1$. Full lines correspond to a linear model, and dashed lines to a polynomial kernel with exponent $\zeta=4$. The GFRE is computed relative to a $\nmax=20$ GTO basis.}
    \label{fig:si-rs}
\end{figure}

These effects can be investigated more easily by considering a 2-body model, that uses only the $\rep<n||\frho_i^1>\propto \rep<n00||\rho_i>$ features.
The comparison between the GTO and the DVR basis (the former being vastly superior in terms of linearly decodable mutual information content\cite{gosc+21mlst}, as seen from the GFRE in the bottom panel of Fig.\ref{fig:si-rs}) is far from clear-cut, with GTOs giving the worst results for forces with $\nmax=4,6$. 
The optimal PCA basis is usually comparable with - but not substantially better than - the best result between GTO and DVRs, for each size of the basis. 
The relative performance of different basis sets is similar when using a linear model and a polynomial kernel, although the nonlinear model reaches an accuracy that is approximately 6 times better for energies and two times better for forces.  
We extend the optimal basis to a PCovR optimization ($\gamma = 0.1$) with the energies as supervised component to determine the contraction coefficients of the basis: as shown in Fig.~\ref{fig:si-rs} (top, center), this PCovR optimal basis yields much better accuracies in the small $\qmax$ range. 
In fact, by taking the ``pure regression'', $\gamma\rightarrow 0$ limit of PCovR, one would obtain a basis that, for a linear model, yields an accuracy comparable to a fully-converged 2-body potential even with $\qmax=1$. 
\rev{
This is because the coefficients are built so that a linear regression performed for the $\qmax$-dimensional features would match as well as possible the predictions of a linear model based on the full primitive basis, 
\begin{equation}
\!\!w^\opt_0 \rep<q\!=\!0; \opt; \gamma\rightarrow 0||\frho_i^1> \!\approx\!
\sum_n w_n \rep<n||\frho_i^1> =  \tilde{y}(A_i).
\end{equation}
Thanks to the spline approximation of the optimal basis, $\rep<0; \opt; \gamma\rightarrow 0||\frho_i^1>$ can be computed at the cost of a single radial function evaluation, much as it would be the case for a pair potential.  
}
The use of a nonlinear model based on the same radial spectrum features provides the simplest test of transferability for the PCovR-optimized basis beyond ridge regression. 
Even though for very small $\qmax$ there is a noticeable improvement (up to a factor of 2 for the force RMSE and $\qmax=2$) against primitive and PCA-optimized bases, the advantage is quickly lost for larger bases, where the variance reduction plays the leading role in driving the selection of radial basis even for small $\alpha$.
As shown in Fig.~\ref{fig:si-rs} (bottom), the improved regression accuracy of PCovR-optimized basis functions comes at a necessary cost in terms of reconstruction error - even though with an intermediate value of the mixing parameter they achieve higher information content than either of the primitive bases, as measured by the GFRE.

\begin{figure}[btp]
    \centering
    \includegraphics[width=1.0\linewidth]{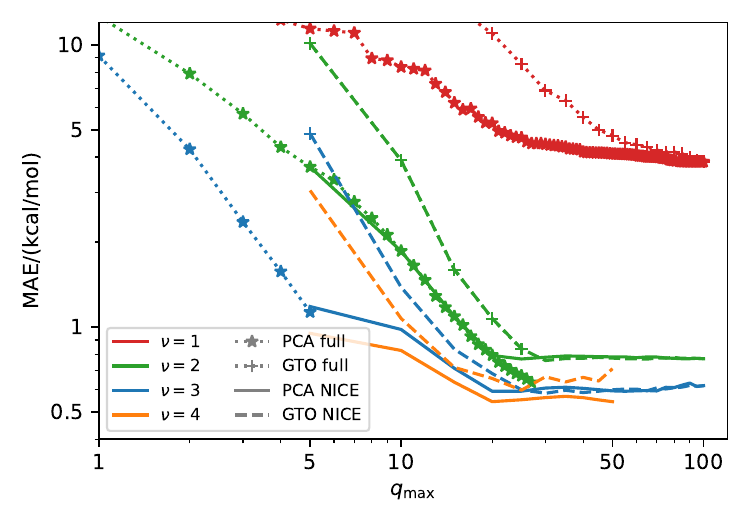}
\includegraphics[width=1.0\linewidth]{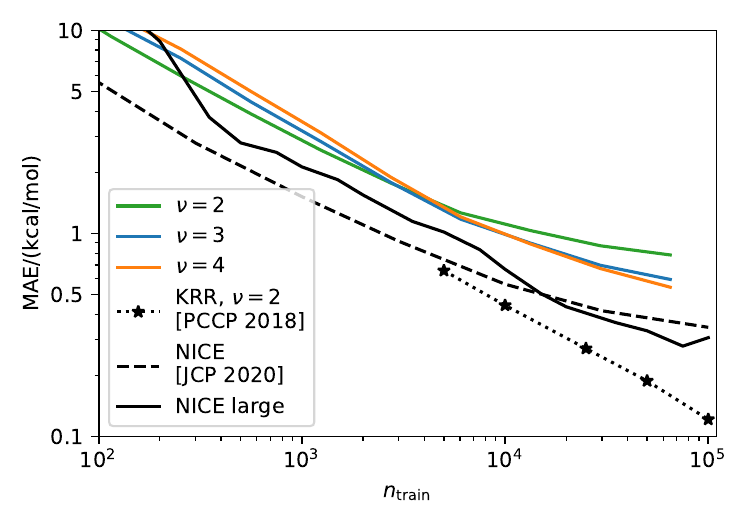}
\caption{Convergence of ML models of the atomization energy of molecules from the QM9 dataset. (top) Convergence as a function of the $(a,n)$ radial basis size, comparing a primitive GTO basis and an optimal PCA contraction, for different body orders of the features. For large $\qmax$ it is necessary to truncate aggressively the NICE iteration, which results in a plateau of the accuracy with large $\qmax$. 
All curves are trained and tested on a set of 65'000 structures, up to the largest $\qmax$ which could fit into 1TB of memory. (bottom) Learning curves are obtained with linear models built on the PCA optimal features of increasing body order. All coloured curves are computed with $\qmax=50$, and the same truncation parameters as in the top panel. For comparison, we show a selection of bespoke models, with black lines: a large NICE model (full line) using 53390 features; the NICE model from Ref.~\citenum{niga+20jcp} (dashed line); a kernel model based on the power spectrum, using parameters analogous to those in Ref.~\citenum{will+18pccp} (dotted line). }
\label{fig:qm9-lc}
\end{figure}

The advantages of using an optimized radial basis become much clearer for the QM9 dataset.  As shown in Fig.~\ref{fig:qm9-lc}, there is a dramatic improvement of performance at all body orders when using a PCA-contracted $(a,n)$ basis, with the improvement becoming more and more substantial for higher $\nu$. 
For the bispectrum features with $\qmax=5$ (effectively only one channel per species), the use of a combined basis leads to a 5-fold reduction of the test error compared to the primitive GTO basis, and makes it possible to reach the symbolic threshold of 1 kcal/mol MAE.
\revtwo{In other terms, an optimal PCA contraction achieves an accuracy comparable to a primitive GTO basis which is roughly 2 times larger. Given that the number of bispectrum ($\nu=3$) features scales as $\qmax^3$, this translates into an order of magnitude improvement in computational efficiency for the QM9 predictions. }
For larger basis sets, and for $\nu>3$, it becomes necessary to truncate the construction of the multispectra, which within the current implementation of the NICE framework is achieved with further PCA contractions applied at each iteration.
In order to be able to use a consistent PCA threshold up to the full primitive GTO basis (which contains $\nmax=20$ radial terms per chemical species) we need to use a rather aggressive truncation, which results in clear performance loss, as evidenced by the saturation of the model accuracy with increasing $\qmax$. 

The interplay of the truncation of the density coefficients, the thresholding heuristic, and the use of the features in a linear or a nonlinear model, is evident in the lower panel of Fig.~\ref{fig:qm9-lc}.
The plot compares the NICE models computed with $\qmax=50$ and an aggressive truncation of the body-order iteration, with the more balanced settings from Ref.~\citenum{niga+20jcp} ($\nmax=12$, $\lmax=7$, $\nu_\text{max}=5$, 1000 invariant features per body order),
with a ``large NICE'' model which includes $53880$ features (up to $\nu=4$, built upon a relatively small spherical expansion  with $\lmax = 5$ and $\nmax=5$) and with a kernel ridge regression (KRR) model that uses the same parameters as in Ref.~\citenum{will+18pccp} (i.e. using only the power spectrum and a nonlinear kernel).
The details of the NICE construction affect substantially the stability and the accuracy of the model in the high-$n_\text{train}$ limit, that vary by a factor of two.
Furthermore, a nonlinear model based on low-body order features is the most accurate, and reaching a state-of-the-art MAE of 0.12kcal/mol with $n_{\text{train}}=10^5$. Even though a thorough investigation of these aspects is beyond the scope of the present work, the understanding of the interplay between the truncation of the density basis and the information loss at higher body order that we discuss here shall support more systematic studies in the future.

\section{Conclusions}

The realisation that most of the widely adopted representations for machine learning of atomistic properties can be seen as a discretization of interatomic correlations naturally points to the importance of determining the most expressive and concise basis to expand the atom density. 
For a given dataset it is possible to  uniquely define a basis that is optimal in terms of its ability to linearly compress the information encoded in the variance of the density coefficients, which can be determined as a contraction of any complete primitive basis, and evaluated efficiently by approximating it with splines. 

We have explored, both analytically and with numerical experiments, the implications of this choice to evaluate higher-order correlations of the density, and to build linear and nonlinear regression models of the energy for both condensed-phase silicon and small organic molecules. 
Our study indicates that the optimization of the density basis has a dramatic impact on the information content of higher-order features, but that
achieving the ultimate accuracy also requires tuning the basis to reflect the sensitivity of the target property to changes in the atomic configurations.
A more intuitive approach may be to perform this tuning at the level of the atomic density, e.g. modulating the amplitude and resolution of atomic contributions depending on the distance from the central atom.
An ``unsupervised'' optimal basis would then provide the most concise, and systematically-convergent, discretization of this tuned atomic density. 

\rev{
Another possible strategy involves the use of supervised criteria in the construction of the basis, as we have demonstrated applying PCovR to the construction of an optimal $\nu=1$ basis. 
A systematic investigation of the effect of varying the parameters of PCovR, as well as the use of PCov-style feature selection\cite{cers+21mlst} in the construction of the multi-spectra, is a promising direction for further research.
One of the challenges is that it is only meaningful to apply the linear reasoning that underlie PCovR to optimize features with the same equivariant properties as the targets, and so the $l>0$ channels of the density coefficients cannot be optimized with a straightforward application of this scheme.
}

\revtwo{ The performance gains associated with the use of an optimal basis are much clearer in the presence of multiple chemical elements, in particular when using a combined basis in which radial channels associated with different species are considered together in the construction of the symmetry-adapted feature covariance matrix. 
This combined basis can capture the same amount of information of a primitive basis that is 3 to 5 times larger, and is essential to the efficient construction of high-order density correlation features, given that we show analytically how the loss of information that is due to a truncated basis becomes worse with increasing $\nu$.} It shall help accelerate the convergence of the schemes, such as NICE, ACE, MTP, that rely on very high body order terms. 
We show that linear NICE models built on high-order combinations of the optimal basis yield much lower error than those constructed on a GTO basis of similar size, even though the truncation of the body order iteration, or introducing nonlinearities, can also affect, positively or negatively, convergence. 

\revtwo{The determination of the optimal basis is much less demanding than the fitting of even the simplest models. After fitting, the evaluation of the contracted basis involves no overhead over a primitive basis of equal size, thanks to the use of a spline approximation.
Given that it provides consistently higher information content, and that it results in models that have comparable (for silicon) or much better (for QM9) accuracy than standard choices of orthogonal bases, we recommend adopting this scheme in any machine-learning approach that requires representing an atomic density -- particularly for systems that involve many chemical species, or for frameworks that rely on the evaluation of high-order density correlations.}

\section*{Data availability}

The data used to perform the calculations presented in this paper are publicly available, and the code to perform the calculations discussed in it can be obtained from public repositories\cite{LIBRASCAL,NICE-REPO}.

\section*{Supplementary material}

The supplementary material contains additional derivations and more detailed benchmarks of the methods discussed in the main text. 

\begin{acknowledgments}
FM, JN and MC acknowledge support by the NCCR MARVEL, funded by the Swiss National Science Foundation (SNSF).
AG, SP and MC acknowledge support from the Swiss National Science Foundation (Project No. 200021-182057).

\end{acknowledgments}

\onecolumngrid
\clearpage
\newpage

\end{document}